# Evolutionary Computational Method of Facial Expression Analysis for Content-based Video Retrieval using 2-Dimensional Cellular Automata

P. Geetha, Dr. Vasumathi Narayanan

**Abstract**— In this paper, Deterministic Cellular Automata (DCA) based video shot classification and retrieval is proposed. The deterministic 2D Cellular automata model captures the human facial expressions, both spontaneous and posed. The determinism stems from the fact that the facial muscle actions are standardized by the encodings of Facial Action Coding System (FACS) and Action Units (AUs). Based on these encodings, we generate the set of evolutionary update rules of the DCA for each facial expression. We consider a Person-Independent Facial Expression Space (PIFES) to analyze the facial expressions based on Partitioned 2D-Cellular Automata which capture the dynamics of facial expressions and classify the shots based on it. Target video shot is retrieved by comparing the similar expression is obtained for the query frame's face with respect to the key faces expressions in the database video. Consecutive key face expressions in the database that are highly similar to the query frame's face, then the key faces are used to generate the set of retrieved video shots from the database. A concrete example of its application which realizes an affective interaction between the computer and the user is proposed. In the affective interaction, the computer can recognize the facial expression of any given video shot. This interaction endows the computer with certain ability to adapt to the user's feedback.

——————————— ◆ ———————————

## 1 INTRODUCTION

A Large amount of video data is publicly available currently. Without appropriate storage and search technique all these multimedia data are not usable. Users want to query the content instead of raw video data. For example, a user will ask for specific part of video, which contains some semantic information. This semantic information extraction plays important role in storing as well as in retrieving process of multimedia data. The process of extracting the semantic content is more complex, because it requires independent person's face and their facial expressions as domain knowledge or user's interaction. Automatic face recognition plays an important role in our society and can be used in a wide range of applications, which could be used to prevent unauthorized access or fraudulent use of ATMs, cellular phones, smart cards, workstations, and Multimedia Information retrieval. However, since faces exhibit significant variations due to illuminations, pose and aging variations, a practical performance of automatic face recognition is difficult. A mature face recognition process could be divided into two steps: feature representation and classification. Development of an automatic facial expression analyzer has attracted great attention in these decades [1].

- *P.Geetha is a Research Scholar, CSE Department, Sathayabama University, Chennai-600109, and India*
- *Dr. Vasumathi Narayanan is a Professor in ECE Department, St.Joseph's College of Engineering, Chennai-600109 and India*

An Automatic Face Analysis (AFA) system is developed in [2] to analyze facial expressions based on both permanent facial features (brows, eyes, mouth) and transient facial features (deepening of facial furrows) in a nearly frontal-view face image sequence. It uses Ekman and Friesen's FACS System [3] to evaluate an expression.

Chandrasiri et al. proposed Personal Facial Expression Space (PFES) to recognize person-specific, primary facial expression image sequences [4]. The key limitation of PFES is that it can not process an unknown face that is not included in the trained person-specific space. In [5] Yeasin et al. used a subjective measurement of the intensity of basic expressions by associating a coefficient for the intensity with the relative image number in the expression image sequence. Though simple and effective for their application, this method does not align expression intensities of different levels. Wang and Ahuja proposed an approach for facial expression decomposition with Higher-Order Singular Value Decomposition (HOSVD) that can model the mapping between persons and expressions [6]. The drawback of their approach is that the global linearity assumption of expression variations introduces some artifacts and blurring while analyzing expressions for persons not contained in the training set. Du and Lin used PCA and linear mapping based on relative parameters as emotional function [7]. They encountered the similar problem as using HOSVD that large amount of train-ing samples are demanded to well represent the variations of expressions for different subjects. Recently Tao et al. proposed general tensor discriminant analysis (GTDA) as a preprocessing step for conventional classifiers to reduce under sample problem [8, 9,10]. How to use this method in facial expression analysis is still an open question. The extraction of image features is one of the fundamental tasks in facial recognition. Recently, a mass of features extraction methods have been reported in literature, including SVD[11], DCT[12], PCA[13,14] and Gabor Features[15,16]. Recently in [17], a new PCA approach called 2DPCA is developed for image feature extraction. Opposite to PCA, the Gabor wavelets have been found to be particularly suitable for image decomposition and representation when the goal is the derivation of local and discriminating features [18, 19, 20]. Belhumeur [21] proposed the FLD algorithm, which could make not only the scatter between classes as large as possible, but the scatter within class as small as possible. In this paper, we use 2DPCA for extracting holistic features and use Gabor wavelets for extracting local features. And then, the FLD has been used for extracting their classifiable features respectively [22]. There many expression analyses done based on stochastic inputs [23]. In this paper, a deterministic approach is proposed to analyze and recognize a facial expression of video shot. One such approach is Cellular automaton (CA) which is a promising computational paradigm that can break through the von Neumann bottleneck [24]. In this paper a 2-dimensional Cellular Automata is considered and each cell is influenced by its neighboring cells including itself. The cells are characterized by its cell states; we consider the states to be either 0 or 1. These influences can be captured through matrices [25, 26], which we call as rule matrices. We first present the model framework, the model specification and evaluation of expression, then the expression analysis and classification for collected video data and finally the store the shot with their identified expression in the database. Different techniques are used to retrieval video content based on query input namely, spatio-temporal formalization methods hidden Markov models and dynamic Bayesian network [27][28][29][30]. For each query input, Video database management systems that allow a user to retrieve the desired video shots among huge amount of video data in an effieicient and semantically meaningful way. A further consideration should be made regarding what it is that the user is looking for.

The paper is organized as follows. In Section 2, describes about 2D DCA and Overview of Proposed Work. In Section 3, describes detailed facial expression analysis and classification based on PIFES is presented. In Section 4, query and retrieval of video shot is discussed. In Section 5, the experiments that have been conducted are presented and discussed. Finally, conclusions and future research directions are presented in Section 6.

## 2 Two-Dimensional Deterministic Cellular Automated and overview of work

### 2.1 2d Deterministic Cellular Automata

This paper addresses 2D Deterministic Cellular Automata for facial expression analysis and recognition based on PIFES from video shots. A cellular automaton is an evolutionary computational method and it is proved by Conway and Wolfram that cellular automata offered promise of computation in a non-directional, parallel manner. It is a discrete dynamic system of simple construction. In a typical cellular automaton, each cell in the array contains one member of a finite set of possible cell values either 0 or 1. A deterministic CA models exhibit a periodic behavior after a certain number of steps. It consists of a grid of cells, usually in one or two dimensions. Each cell takes on one of a set of finite, discrete values. For concreteness, in this paper we shall refer to two-dimensional grids, each cell has a finite and fixed set of neighbors, called its neighborhood. In this paper, we Used Von Neumann neighborhood. In von Neumann neighborhood, each cell has neighbors to the north, south, east and west; In general, in a d-dimensional space, a cell's von Neumann neighborhood will contain 2d cells. Each cell follows the same update rule, and all cells' contents are updated simultaneously. A critical characteristic of CAs is that the update rule examines only its neighboring cells so its processing is entirely local; no global or macro grid characteristics are computed. These generations proceed with all the cells updating at once. The global behavior of a CA is strongly influenced by its update rule. We showed here the three fundamental properties of cellular automata:

*Parallelism* : A system is said to be parallel when its constituents evolve simultaneously and independently. In that case cells updates are performed independently of each other.

*Locality* : The new state of a cell only depends on its actual state and on the neighbourhood.

*Homogeneity* : The laws are universal, that's to say common to the whole space of cellular automata.

In this paper, we proposed the combination of all the three properties of CA. i.e., Multiple CAs are generated for each region of face i.e eye brows, eye lids, eyes, lips, mouths, cheeks and head to find the universal expression of any frame sequence of video shot. Since to generate the rule matrix of each facial region simultaneously, we used Locality property and it is done for each facial region in parallel by using Parallelism property of CA. We generated universal facial expressions which could be done using Homogeneity property.

## 2.2 Overview of Proposed Work

This paper presents novel classification algorithms for recognizing dominant object facial activity using cellular automata. This requires the following preprocessing step.

1. Input movie video

2. Segment this video into number of scenes which in turn segmented into shots using a tool.

3. Create key database based on PIFES which holds all different expressions of individual persons.

The system framework of Cellular Automata based Video Shot Classification is given in Figure1. After the preprocessing step, the proposed method has four phases. They are Facial Expression evaluation by Cellular Automata, Facial expression analysis which involved face detection, Key face Extraction and face recognition, Expression Classification phase and finally, Query Input given by the user and retrieval of relevant shots is kaken place based on similar expressions. The following system framework of Figure1 explains clearly the possible input and output in each stage of proposed work.

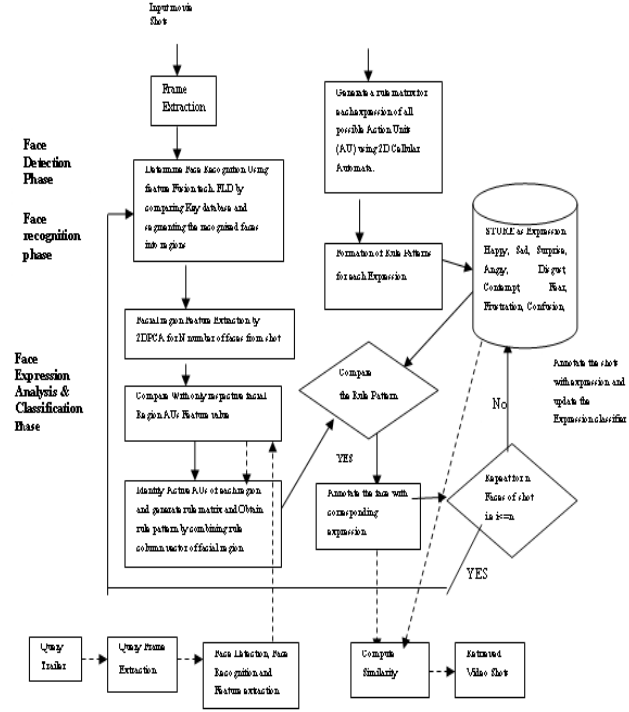

Fig1: Overview of Proposed Work

In first phase, a distinctive rule is generated for different kind of expressions for each section of face by 2D Deterministic Cellular Automata. This is done by combining possible AUs of different facial regions. Then rules are updated synchronously. Finally determine the number of active cells in a rule matrix to annotate an expression for the respective combinations of AUs and optimize it. Then it is stored in a database. In second phase, before analyzing an expression of given set of faces in a shot, it involves sub sections. They are face detection, key face extraction and face recognition. In this face detection sub section, frames must be extracted from a video shot and then face detection is done by combination of RGB, YCbCr and HIS skin colors algorithms. Key faces are extracted from detected set by computing Eigen values. Then it is to recognize the key faces two dimensional principle component analysis and Gabor wavelets and then fuses these features with Fisher's Linear Discriminant (FLD). This fused FLD feature value is compared with fused FLD feature value of train database. If there is match with database then those faces are only leads to expression analysis. To compute an expression for each key faces, a feature value is extracted by using 2DPCA technique for each facial region of it, and also computes the same for AUs of FACs and generate rule pattern.

In third phase, for any number of key faces of shot, an expression is evaluated by comparing with each expression's rule pattern in the database. If there is match then annotate them according to an expression and stored in database. In the fourth phase, query trailer clip is given as input; expression evaluation step i.e phase 2 is repeated in this phase also to identify an expression of query frame. This expression of query frame is compared with the database content. If there is a match then relevant shots with the same expression are retrieved and played to the user. This is represented as dotted line in Figure1.

## 3. Detailed Facial Expression analysis and Classification based on PIFES

THIS SECTION EXPLAINS IN DETAIL ABOUT HOW THE FACIAL EXPRESSION OF ANY PERSON IS ANALYZED AND CLASSIFIED IN THE DATABASE.

### 3.1 Facial Action Units

In 1978, Paul Ekman and Wallace V. Friesen published the Facial Action Coding System (FACS) [3], which, 30 years later, is still the most widely used method available. Through observational and electromyographic study of facial behavior, they determined how the contraction of each facial muscle, both singly and in union with other muscles, changes the appearance of the face. Rather than using the names of the active muscles, FACS measures these changes in appearance using units called Action Units (AUs). FACS is an efficient, objective method to describe facial expressions. *Figure 2* illustrates some of these Action Units and the appearance changes they describe. The benefits of using AUs are two-fold. First, individually and in combination they provide a way to unambiguously describe nearly all possible facial actions. Second, combinations of AUs refer to emotion-specified facial expressions. For instance, an expression is distinguished by the combination of primary and auxiliary AUs which are tabulated in Table 1.

| Expressions | AUs |
|---|---|
| Happiness | 6 + {12+16+25+26} |
| Sadness | {1+4}+7+{15+25+28}+63 |
| Angry | {2+4}+7+{16+23+24+25+26} |
| Disgust | 10+61 |
| Fear | {1+4}+{5+7}+{20+25+26} |
| Surprise | {1+2}+5+{26+27} |
| Contempt | 4+6+{10+24} |
| Frustration | 2+28+{43+64} |
| Confusion | 1+5+25 |

Table 1: Distinctive Combinations of AUs for different Expression

### 3.2 Definition of Cellular Automata as applied to Facial Expression Learning

Let M denote the set of Action units developed by by Ekman and Friesen in their FACS [3]. A 2-D cellular space is a 4-tuple,(Z, S, N, tf), where Z is a Lattice of M X *M* where M varies as {1,2,3,4,5,6) integer value depending on the facial region such as Cheeks, eyes lids, eye brows, eyes, lippart1, lip part2 of AU sets respectively, S is a set of cellular states, most commonly the number of states is 2. In one of the states the cells are often said to be active denoted as **1**, in the other to be inactive denoted as **0**. *N* is the type of neighborhood e.g. n neighbors, and t*f* is the local transition function. An application of transition rules is that one generation produces another generation and then is applied again, and so on. The relevant neighborhood function is a function from M$xM$ into $2^{MxM}$ defined by h($\alpha$)={$\alpha+\delta 1$, $\alpha+\delta 2$,…, $\alpha+\delta n$}, for all $\alpha$ ε $MxM$, where $\delta_i(i = 1,2,…,n)$ ε $MxM$ is fixed. The neighborhood state function of a cell $\alpha$ at time $t$ is defined by $h^t(\alpha)=(S^t(\alpha+\delta 1),S^t(\alpha+\delta 2),…,S^t(\alpha+\delta n))$.

We want to find some algebraic properties $2^9$ = 512 linear CA rules that are quite useful to describe the behavior of M x M of 2-D Cellular Automata where M=3. Rule 1 characterizes dependency of the central cell on itself alone whereas such dependency on it's either of four neighbors is characterized by rule 2,8,32 and128. These are the five fundamental rules are Rule1, Rule2, Rule4, Rule 8 and Rule 16 and other fundamental rules are Rule32, Rule64, Rule128 and Rule 256 can be obtained from these with a transpose operation. It is represented as below:

| $2^6$ | $2^7$ | $2^8$ |
|---|---|---|
| $2^5$ | $2^0$ | $2^1$ |
| $2^4$ | $2^3$ | $2^2$ |

For example, the 2D CA rule 1 either cell value (0+0+0+0+1) refers to the four neighborhood dependency of the central cell on top, left, bottom, right or itself. The number of such rules is $^9C_0 + {}^9C_1 + … + {}^9C_9$ =512 which includes rule characterizing no dependency. Also for this problem matrix is a matrix of a fixed dimension with respect to space of locally connected cells whose mutual interactions determine their behaviors. For eg. A problem matrix of dimension (3 x 3) and rule matrix is generated with the dimension of (9 x 9) in which Rule 1 is represented as (9 x 1) column vector [0 0 0 0 0 0 0 0 1]$^T$ ). In this set of rule matrix result which would have many unwanted rules and they are deleted by checking with self looping of action units. The "update rule" to determine what the next cell in the evolution will be. Typical elementary update rules used in this work are of the form, i.e Rule 1 (0 + 0 + 0 + 0 +1) is applied uniformly to cell of a problem matrix with null boundary. If so, update the either northeast of southwest cell be active. This process of framing rule matrix and optimization is done for Action units of each facial region i.e eye brow, eyelids, eyes, cheeks and mouth in parallel.

### 3.3 Learning of Expressions using FACs

In this section, we explained about the FACs and the rule generation using 2D deterministic cellular automata.

### 3.3.1 Expression Evaluation using Cellular automata and FACs

The Automated Facial Expression Recognition Classification System for Independent persons is proposed in this paper. We have Automated FACs using 2D-DCA (deterministic cellular automata) and this portable, near real-time system will detect the TEN expressions of emotion including neutral, as shown in figure2, from successive Frames of scenes and classify them according to their expressions.

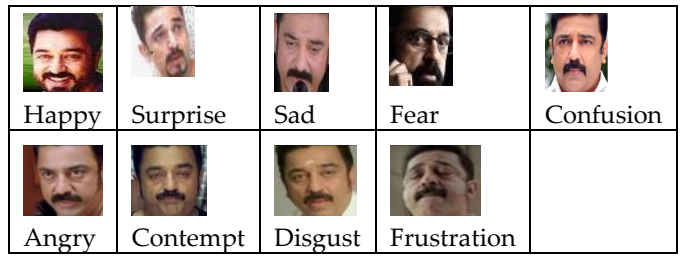

| | | | | |
|---|---|---|---|---|
| Happy | Surprise | Sad | Fear | Confusion |
| Angry | Contempt | Disgust | Frustration | |

Figure2: Demonstrates the Ten Expressions.

### 3.3.2 Formation of Rule pattern for Individual facial Region

Here we had chosen around 20 Aus for expression evaluation. First of all, we group the AUs based on facial regions such as cheeks, eyes lids, eye brows, eyes, lippart1, and lip part2. Table 3 lists the facial region along with their AUs.

| Facial Region | AU List | Value for M in CA |
|---|---|---|
| Cheeks | 6 | 1 |
| Eyes Lids | 5,7 | 2 |
| Eye Brows | 1,2,4 | 3 |
| Eyes | 43,61,63,64 | 4 |
| Lip Part1 | 10,16,25,26,27 | 5 |
| Lip Part2 | 12,15,20,23,24,28 | 6 |

Table3: List of AUs with M value

We would give an example to explain how rule patterns are generated for each expression. Deterministic rule pattern is evolved using 2D Cellular Automata for the set of AUs published by Paul Ekman and Wallace V. Friesen as Facial Action Coding System (FACS). For example, the rule pattern for eye brows face region is evolved using 2D CA as { Z,S,N,tf} where Z is M*M, M=3 in this case and S={0,1}, N=5 neighborhoods. finally the transition function is defined as

"if current cell is active ie. in red colour and top, left, right and bottom of center cell or a pair {top, left} or {bottom, right} are dead ie in black colour then either southwest or northeast cell become active and center cell become dead"

Consider an example, originally all cells are assumed to be dead. To show the Rule pattern for AU=2 are given below:

| AUs | 1 | 2 | 4 |
|---|---|---|---|
| 1 | 0 | 0 | 0 |
| 2 | 0 | 1 | 0 |
| 4 | 0 | 0 | 0 |

Rule column vector is evolved as $\{0\ 0\ 0\ 0\ 1\ 0\ 0\ 0\ 0\}^T$ and apply the update rule to this set to evolve next generated rule vector. Cellular automate is defined as

| AUs | 1 | 2 | 4 |
|---|---|---|---|
| 1 | 1 | 0 | 0 |
| 2 | 0 | 0 | 0 |
| 4 | 0 | 0 | 0 |

Rule column vector is evolved as $\{1\ 0\ 0\ 0\ 0\ 0\ 0\ 0\ 0\}^T$ to set AU value as 1. This step used for evolving distinctive rule pattern for each Action Unit. The configuration it yields on successive time steps is thus simply obtained with respect of Active units of facial regions.

Distinctive set of 9*9 rule matrix is evolved using this evolutionary method and out of them only fundamental rules are extracted such that there are the five fundamental rules are Rule1, Rule2, Rule4, Rule 8 and Rule 16 and other fundamental rules are Rule32, Rule64, Rule128 and Rule 256 can be obtained from these with a transpose operation. In this fundamental rules set, only diagonal active rule are stored in a facial region rule database. Finally we stored only three rule vector instead of 81 rule vector. These processes are repeated for each facial region action unit set of FACs.

| AUs | Rule vector |
|---|---|
| 1 | 100 000 000 |
| 2 | 000 010 000 |
| 4 | 000 000 001 |

### 3.3.3 Expression Evaluation

All expressions are evaluated by concatenating rule column vector of each facial region in sequence and each of them are separated by special symbol $ in order to distinguish and store the same in the database. For eg: following state transition explains how the rules of various facial regions are concatenating their rule vector in sequence.

Start

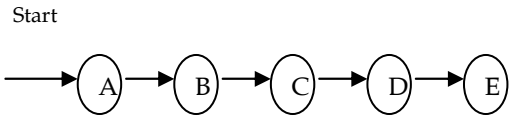

**State Name: A-Eye Lids, B-Eye brows, C- Eyes, D- Cheeks**

**E-LipPart1/LipPart2**

It is assumed that all facial region action units need not be exist in a training frame to analyze its expression. It means that neutral expression of corresponding facial region is concatenated to form a rule pattern. Similarity, many rule patterns for each expression might be raised and every possible combination of rule pattern of various facial regions are extracted from the existing set of action unit in FACs.

Consider an example, for a Happy Expression, as in table 1, only cheeks and lip related action units are involved and others are in neutral in expression. Hence rule pattern1 for the set of involved only action unit {0, 0, 0, 6, 12}, and here 0 in the set interprets that except cheeks and lips, all others are in neutral in expression, is {00$000$0000$1$100000$}. Here lip part2 holds 6*6 rule matrixes which are enough to represent as 100000 remaining bit positions are padded with zeros only. Generating distinctive combinations of rule pattern for various expression using deterministic cellular automata and store the same in the database for analyzing forthcoming video shots.

### 3.2. Expression Analysis for Training Input

This section should be executed in sequence to analyze facial expression based on PIFES for any test video.

### 3.2.1. Frame Extraction from Video shot

The video processing component of the DCA-FEA application is responsible for sequencing the inputted video into individual scenes using frame grabbing method at a rate of 25 frames per second. Once the video is sequenced, the shots of frames place onto a queue for input into the Expression Analysis Module.

### 3.2.2. Face Detection

The existence and count of faces present in a given set of frames of shots are evaluated by using combined skin color algorithm [10]. This must be evaluated for any kind of head position shown in figure3. By combining the detected regions from all the three algorithms, skin region is extracted. Sample different head position of any person in that movie frame was extracted by using this combined skin algorithm is shown in figure4. First, extract the facial features and drawing bounding box around the face region. The skin region is extracted by an assumption that if the skin frame is detected by one or more algorithm(s) and for the same frame other algorithm gives the false result, then also the face is extracted using the combination algorithm. In order to detect the faces, after getting the skin region, facial features i.e., Eyes and Mouth are extracted. The frames are obtained after applying skin color statistics is subjected to transformed to gray-scale frame and then to a binary image by applying suitable threshold. This is done to eliminate the hue and saturation values and consider only the luminance part. This luminance part is then transformed to binary image with some threshold because the features we want to consider further for face extraction are darker than the background colors.

### 3.2.3 Face Recognition

In this work, set of key face database of different person is maintained. After the detection of faces from given set of frames, we calculate the Eigen faces from the training set, keeping only the M Eigen faces which correspond to the highest Eigenvalues in order to speed up the recognition process. This is determined by computing the difference between mean image of key face database and training set. These M images denote the "face space". Finally we calculate the corresponding location or distribution in M-dimensional weight space for each known individual, by projecting their face images (from the training set) onto the "face space". When a new face image is encountered, we calculate a set of weights based on the input image and the M Eigenfaces by projecting the input image onto each of the Eigenfaces. We determine if the face in a frame is a face (known or unknown) by checking to see if the image is sufficiently close to "face space" — i.e. determining the ability of the Eigenfaces to reconstruct the image. If it is a face, we classify the weight pattern as either a known person or as unknown person [21].

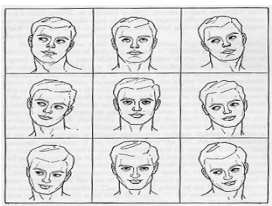

Figure3: Possible Head position Frame set

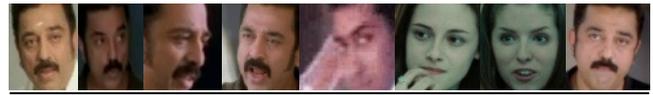

Figure4: Sample independent Person head Position

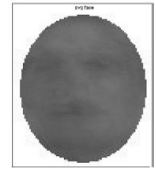

Figure5: Average mean face from of human

To reduce the dimensionality of the Eigen vector space and obtain more useful features for subsequent pattern discrimination, the FLD technique is used here [21]. The above processed Eigen faces are used as the training set for the face recognition. To recognize face, use Euclidean distance measurement with the key face database and is defined as in Eq1.

$$||E\_dis|| = \sqrt{||\varepsilon_i - \mu_i||^2} \quad \text{------ (1)}$$

Find $er = \min(||E\_dis||)$ and if $er < \Phi$, where $\Phi$ is a threshold chosen heuristically, then we can say that the training face is recognized as the key face with which it gives the lowest score and store these training faces of shot in the database for further expression evaluation. If $er > \Phi$ however then the training face does not belong to the database and ignore it.

Next step is used to evaluate the facial expression analysis and classification.

### 3.2.4. Expression Analysis and Classification

Each image is represented in two ways. The first uses fiducial points which could be selected manually or automatically. A second way is to use any feature extracting technique for extracting feature value from the face. In this paper, we used 2DPCA technique for feature extraction. For each Eigen face from training set, sections of facial regions are obtained as Eyes, Cheeks and Lip and also we determine 2DPCA feature value in sequences of section. We have to determine the same featured value for each example images listed in FACs. To recognize an equivalent action unit, it is enough to find the similarity distance measurement, such as Euclidean distance, between feature values of corresponding Example image in FACs and sequenced sections of training face. It is then used to retrieve the corresponding facial region action unit with very small error %. This process is repeated for each section of facial regions in sequence and concatenates the rule vectors to form a final rule pattern which in turn compared with rule pattern of expression in a database. If there is a match then the first face of frame of shot is annotated as 1 and if second face is identified as different expression then it is annotated as 0 or 1 only. This expression analysis is repeated for as many faces of frames as present in a shot and finally concluding shot expression as majority facial expression of frames by taking log-ic OR operation on annotation. Then the corresponding shot is added with that frame sequences in the expression pool. This expression analysis and classification is repeated for any number of scenes in a movie and for multiple movies. Next step gives query phase in detail.

## 4 Video Shot Retrieval Phase

In order to retrieve a relevant shot, a query trailer is taken in which a user should select a frame. After it undergoes the steps as per mention in section 3.2, an expression of that query frame is analysed which in turn generate the rule pattern. Then compute the similarity of query input face expression's rule pattern with the database content. If there is a match then the relevant shots are retrieved and played the same to the user.

## 5 Experiments Results

### 5.1 Data set used
*Sequences of frames extracted from video scenes of single movie or multiple movies.
*Facial behavior in 100 adults –female and male of ages between 10 and 45 years was recorded.
*Subjects sat directly in front of the camera and performed a series of facial expressions that included single action units.

### 5.2 Sample Output
Each expression is identified from a neutral face. Each frame in the sequence was digitized into a 375 by 300 pixel array with 8-bit precision for gray scale values. For each expression, action units were coded by 2D-DCA as explained in section 3.1.1. Action units are important to the communication of emotion and which are selected for analysis. This frequency criterion ensured sufficient data for analyzing and classify a facial expression based on action unit occurred in combination with other action units.

For facial-feature tracking with PCA, we used 800 samples of more than 20 action units or action unit combinations that occurred image sequences of movie. The samples were randomly divided into training and cross-validation sets. However, if for a subject, action unit or combination of action unit is occurred in it then rule pattern is comparison taken place and thus that what was recognized by our method was the rule pattern of action unit rather than the subject.

Finally, we present an example of facial expression understanding in a live image sequence. The original image sequence of facial expressions has 500 frames containing the ten emotional expressions plus the neutral state among them. We provide selected frame sequences that will, hopefully, convey our results.

In this paper, we chosen Tamil movie is taken as an input movie. In this movie, shots are segmented by using Total Video Converter tool and faces with different expressions are stored in a database before processing. By inputting shots, the experiments show very good results for combined skin color algorithm. The results obtained using the previous conditional probabilities and threshold value. Total number of frames is present in that shot1, of size 2.43 MB, is 143 and existence and count of faces in that frame are estimated by an algorithm very accurately. The accuracy is found to be 96.59%. Sample results from proposed algorithm are shown in Table2.

| Name of Movie (No. of Scenes) | Average No. of Frames per Shots | Face Detection- Only RGB color space Overall Accuracy % | Face Detection- By Skin Color Combined (RGB +YCbCr + HIS) Alg. - Overall Accuracy % | Face Recognition- Only FLD- Overall Accuracy % | Face Recognition- Feature Fusion Algorithm(2 DPCA + Gabor + FLD)- Overall Accuracy % |
|---|---|---|---|---|---|
| Movie1 (15) | 145 | 55% | 97% | 88.85% | 96.59% |
| Movie2 (15) | 150 | | | | |
| Movie3 (16) | 123 | | | | |
| Movie4 (14) | 114 | | | | |
| Movie5 (15) | 153 | | | | |
| Movie6 (15) | 115 | | | | |
| ⋮ | | | | | |
| Movie14 (14) | 124 | | | | |
| Movie15 (14) | 164 | | | | |

**Table2:** Sample results of Combined Skin Color face detection algorithm and Face recognition with Key face database.

Sample output this automatic facial expression analysis and classification is given in detail in Figure6. The feature values of each section are matched with the set of AUs and retrieve them. Further to refine the AUs set, rule pattern is generated for each sections of training faces, which is then compared with rule pattern of Example image group. Categorization of AUs can be used for the classification of various expressions. To retrieve a relevant shot with respect to the query input's expression should taken place in the same manner as we explained in section 3.2. Then compute the pattern match between the expressioj of query input and the database content. If there is a match then the relevant shot is retrieved and played the same to the user. The sample screen shot of querying and retrieving relevant shot from the database is shown in Figure7.

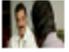

Figure6 Expression Identification from the sequence of Frames

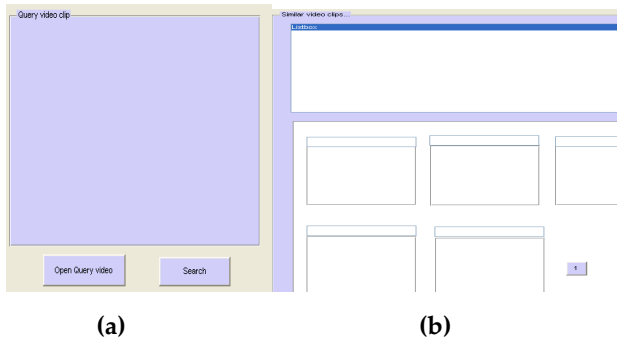

(a)          (b)

Figure 7: (a) Query Search Screen shot (b) Retrieval of relevant shot based on similar expression

### 5.3 Performance Analysis

We conducted this expression evaluation algorithm for more than 10 movies. In classification, the Precision and Recall are defined as the terms true positives, true negatives, false positives and false negatives, are used to compare the given classification of an item (the class label assigned to the item by a classifier) with the desired correct classification (the class the item actually belongs to) and Accuracy for a group are computed. Usually, Precision and Recall scores are not discussed in isolation. Instead, either values for one measure are compared for a fixed level at the other measure or both are combined into a single measure, such as the F-measure, which is the *weighted harmonic mean of precision and recall* and in our work its value is 96.1% .Table3 gives the comparison of other Facial expression classification with our work and it shows that our work would gives the better recognition accuracy rate than the existing methods and also shown the same comparison as chart in Figure7 and Precion-Recall Comparison chart is shown in Figure8.

| Facial Expression Recognition Methods | The average Recognition accuracy Rate (%) |
|---|---|
| Neural Networks (SVM)-Still Image | 86.00% |
| Neural Networks (HMM)-Still Image | 90.10% |
| Dynamic 2D Cellular Automata-Video Frames Expression Classifier | 94.13% |

Table3: Recognition accuracy of various facial expressions Methods with Proposed Work

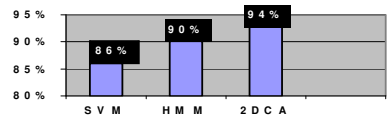

Figure7 Comparison Existing method with Our Work

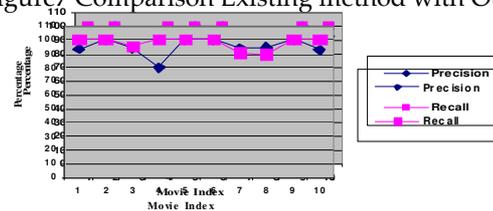

**Figure8 Precision-Recall Comparison of Our Proposed Work**

## 6 CONCLUSIONS

In recent years, human facial expression analysis has become an active research area. Various approaches have been made towards robust facial expression recognition, applying different image acquisition, analysis and classification methods. Facial expression analysis for a video shot is an inherently multi-disciplinary field and it is important to look at it from all domains involve in order to gain insight on how to build reliable automated facial expression analysis system. In this paper, first coding facial expression with an appearance-based representation scheme such as FACs and then using deterministic 2D Cellular Automata in order to translate recognized facial expression both spontaneous and posed. It has been developed in this paper as general mathematical models. This paper thus proposes a Person-Independent Facial Expression Space (PIFES) to analyze and synthesize facial expressions based on Partitioned 2D-Cellular Automata which capture the dynamics of facial expressions and classify the video shots based on it. T Target video shot is retrieved by comparing the similar expression is obtained for the query frame with respect to the key frames in the database video. Consecutive key frames expression in the database that are highly similar to the query frames, then the key frames are used to generate the set of retrieved video shots from the database. A concrete example of its application which realizes an affective interaction between the computer and the user is proposed. In the affective interaction, the computer can recognize the facial expression of any given video shot. This interaction endows the computer with certain ability to adapt to the user's feedback. The proposed model could be incorporated in global emotion recognition systems, including others elements recognition, such as voice intonation or concentration.

&Technology.